# Brain-inspired Chaotic Graph Backpropagation for Large-scale Combinatorial Optimization

Peng Tao[1], Kazuyuki Aihara[2], Luonan Chen[1,3,*]

[1] *Key Laboratory of Systems Health Science of Zhejiang Province, School of Life Science, Hangzhou Institute for Advanced Study, University of Chinese Academy of Sciences, Chinese Academy of Sciences, Hangzhou 310024, China.*

[2] *International Research Center for Neurointelligence, The University of Tokyo Institutes for Advanced Study, The University of Tokyo, Tokyo 113-0033, Japan.*

[3] *Key Laboratory of Systems Biology, Shanghai Institute of Biochemistry and Cell Biology, Center for Excellence in Molecular Cell Science, Chinese Academy of Sciences, Shanghai 200031, China.*

\* To whom correspondence may be addressed: Luonan Chen: lnchen@sibs.ac.cn

# Abstract

Graph neural networks (GNNs) with unsupervised learning can solve large-scale combinatorial optimization problems (COPs) with efficient time complexity, making them versatile for various applications. However, since this method maps the combinatorial optimization problem to the training process of a graph neural network, and the current mainstream backpropagation-based training algorithms are prone to fall into local minima, the optimization performance is still inferior to the current state-of-the-art (SOTA) COP methods. To address this issue, inspired by possibly chaotic dynamics of real brain learning, we introduce a chaotic training algorithm, i.e. chaotic graph backpropagation (CGBP), which introduces a local loss function in GNN that makes the training process not only chaotic but also highly efficient. Different from existing methods, we show that the global ergodicity and pseudo-randomness of such chaotic dynamics enable CGBP to learn each optimal GNN effectively and globally, thus solving the COP efficiently. We have applied CGBP to solve various COPs, such as the maximum independent set, maximum cut, and graph coloring. Results on several large-scale benchmark datasets showcase that CGBP can outperform not only existing GNN algorithms but also SOTA methods. In addition to solving large-scale COPs, CGBP as a universal learning algorithm for GNNs, i.e. as a plug-in unit, can be easily integrated into any existing method for improving the performance.

# 1  Introduction

The goal of combinatorial optimization problems (COPs) is to find the solution corresponding to the smallest objective function among a limited but often a large number of candidates, which is closely related to many hard problems in science and industry, such as logistics and transportation, circuit design and drug development[1-5]. During the long research history of combinatorial optimization, many classical methods have been proposed, such as branch-and-bound and Tabu search methods, but with the advent of the era of big data, the scale of real-world COPs is often very large, and traditional methods usually have difficulty in obtaining high-quality feasible solutions in an acceptable time. Therefore, it is of great theoretical and practical importance to develop new generalized solution methods suitable for large-scale COPs.

In recent years, new combinatorial optimization algorithms different from traditional operations research-based methods have been developed. Among them, machine learning-based methods are currently a hot topic in research. This type of algorithm can be further divided into three subclasses based on the learning paradigm, namely supervised learning, reinforcement learning, and unsupervised learning. Early combinatorial optimization methods mainly used supervised learning, which minimizes some empirical loss functions to complete the complex, nonlinear mapping from the input representation of a problem to a target solution. For example, the classic Pointer network (Ptr-Net)[6] uses a Seq2Seq model to generate variable-length permutation combinations and then solve the traveling salesman problem (TSP). However, the feasibility and performance of supervised learning-based methods largely depend on the availability of a large amount of pre-solved high-quality training data, which is often difficult to obtain[7]. In contrast, reinforcement learning techniques aim to learn a policy that maximizes some expected reward function and can therefore bypass the need for training labels. Specifically, COPs can usually be described by an objective function, which can be used as the reward function in reinforcement learning. For instance, Bello et al.[8] constructed an actor-critic model by combining a Ptr-Net-based decoder with a recurrent neural network (RNN) encoder and using the expected length of a tour as the reward signal, which served as an approximate solver for the TSP. Khali et al.[9] further combined reinforcement learning and graph embedding to gradually construct feasible solutions through learning an efficient greedy meta-heuristic method, which achieves decent results in problems like minimum vertex cover (MVC), maximum cut (MC) and TSP. However, reinforcement learning-based methods often require manual modeling of the target problem, such as defining states and actions, which typically demands rich prior knowledge. Consequently, some works have attempted to use reinforcement learning to assist traditional combinatorial optimization methods[5]. For instance, NeuroLKH[10] uses reinforcement learning to assist the Lin-Kernighan-Helsgaun (LKH)[11] algorithm in selecting candidate sets of edges, thereby improving the quality of the TSP solution. Nonetheless, strategies learned in this manner are difficult to extend from one COP class to another. To address the above issues, Schuetz et al. recently proposed a more general framework, called physics-inspired graph neural network (PI-GNN)[12], which transforms combinatorial optimization problems into quadratic unconstrained binary optimization (QUBO) problems. The corresponding Hamiltonians of these problems are encoded as the loss functions of GNNs, and the GNNs are then trained in an unsupervised manner to obtain the solution to the corresponding COP. It should be noted that although PI-GNN has advantages such as high computational efficiency (as it can solve combinatorial optimization problems of up to one million variables[12]) and strong scalability, the quality of its solutions is not superior to that of traditional state-of-the-art (SOTA) methods. The reason for this issue is that the solution of PI-GNN depends significantly on the performance of the GNN learning algorithm. Currently, GNN learning still uses gradient dynamics-based backpropagation (BP) algorithms and its variants, such as Adam[13], thus making GNN models prone to local minima, which limits the quality of the solution. Therefore, developing a new gradient-free dynamic universal GNN learning algorithm to overcome the local minima problem is of great importance for solving large-scale combinatorial optimization problems.

On the other hand, chaotic dynamics is one of the most important theoretical discoveries in natural sciences in the 20th century, which describes unpredictable global dynamics generated by deterministic systems[14]. Specifically, chaotic dynamics is highly sensitive to initial conditions, and exhibits randomness and unpredictability in the long-term evolution. Yet since its dynamics is fundamentally deterministic, this randomness is known as pseudo-randomness[14]. Moreover, theoretical studies have shown that chaos has global exploratory properties due to the ergodicity in a fractal space of a strange attractor[15]. It is due to these dynamic properties that chaotic dynamics has been widely used to solve global optimization problems[15-19]. In addition, chaotic dynamics has been observed in the heart[20], neurons[21, 22], and even the brain[23], and it is very important for many biological processes[24], such as gene expression and regulation[25], signal processing[26], and sleep[27]. Among them, Skarda and Freeman's experiments showed that the dynamics of the rabbit brain is chaotic at the resting state and works for learning new odor patterns[23]. Matsumoto et al. found that squid giant axons show chaotic responses to periodic external stimulation[28]. Moreover, recent experiments have also found that the neural networks of animal brains are in a critical or quasi-critical state between order and chaos[29, 30]. These studies imply a possibility that over long periods of evolution, animal brains have been able to use chaotic dynamics for cognition or learning. Therefore, is it possible to draw on the chaotic dynamics of the brain to establish a new, universal GNN learning theory and method?

Based on the above motivation, we developed the chaotic graph backpropagation (CGBP) algorithm in this work (Fig. 1). By introducing a local loss function to simulate chaotic dynamics in the brain and utilizing its pseudo-randomness and global ergodicity, CGBP overcomes the drawbacks that existing BP-based GNN learning algorithms are prone to local minima, and achieves high-precision results in large-scale COPs, outperforming not only the existing GNN learning algorithms but also SOTA methods. In addition to solving COPs, CGBP can be used as a universal learning algorithm for training GNNs, i.e. it can be easily integrated into any existing method as a plug-in unit for improving the performance. From both computational and theoretical viewpoints, we can show that there is indeed chaotic dynamics in CGBP, i.e. Marotto chaos[31] generated from a snapback repeller when hyperparameter $z$ is sufficiently large. In the following, we will introduce the CGBP algorithm in detail and then compare it with existing SOTA methods on large-scale benchmark COP datasets.

## 2 Chaotic graph backpropagation

### 2.1 Combinatorial optimization

Typical combinatorial optimization problems include TSP, MC, MVC, maximum independent set (MIS) and graph coloring (GC). Assuming that the number of variables in a combinatorial optimization problem (COP) is $n$, a feasible solution can be represented as a vector $\mathbf{x} = (x_1, x_2, \cdots, x_n)$. The objective of a COP is to find the feasible solution with the smallest objective function value.

When the state of each variable is binary ($x_i \in \{0,1\}$), many COPs can be represented under the QUBO[32] framework, such as MIS and MC. At this time, the COP corresponds to the Ising model in physics[33], and its objective function is equivalent to the Hamiltonian of the Ising model, given by

$$H = \mathbf{x}^T Q \mathbf{x} = \sum_{i,j} x_i Q_{ij} x_j, \qquad (1)$$

where $Q$ is a constant matrix encoding the COP. For example, for the MIS problem, the objective is to find a subset of nodes $\mathcal{V}_s \subseteq \mathcal{V}$ from a given graph (denoted as

$\mathcal{G}(\mathcal{V}, \mathcal{E})$, where $\mathcal{V}$ and $\mathcal{E}$ are the sets of nodes and edges, respectively) such that the number of nodes in $\mathcal{V}_s$ is maximized under the premise that there is no edge between any two nodes in $\mathcal{V}_s$, and the corresponding Hamiltonian is

$$H_{MIS} = -\sum_{i \in \mathcal{V}} x_i + P \sum_{(i,j) \in \mathcal{E}} x_i x_j, \tag{2}$$

where $P$ is a penalty parameter, usually set to 2.

When the number of the state is higher than 2, these COPs correspond to the Potts model (a generalization of the Ising model), and the form of its Hamiltonian is similar to that in Eq. (1). At this time, it is only necessary to perform one-hot encoding on the variable $x_i$, and this will be further introduced in the Method section.

## 2.2 Graph neural networks

Graph neural networks (GNNs) are neural networks that can aggregate and represent information on graphs. Typically, a GNN layer consists of three functional units: message passing, aggregation and nonlinear activation. When multiple GNN layers are stacked, the feature information of nodes can be propagated along edges on the graph, and the greater the number of layers, the wider the range of information propagation. For example, the computation process of a graph convolutional network (GCN)[34] can be represented by the following updating formula

$$h_i^{(l+1)} = \sigma\left(\sum_{j \in \mathcal{N}(i)} \frac{1}{c_{ij}} h_i^{(l)} W^{(l)} + B^{(l)}\right), \tag{3}$$

where $h_i^{(l)}$ represents the feature vector of the $i$th node in the $l$th layer, $W^{(l)}$ and $B^{(l)}$ represent the weights and biases in the $l$th layer, $\mathcal{N}(i)$ represents the set of neighbors of node $i$. $c_{ij}$ is the production of the square root of the degree of node $i$ and $j$, namely $\sqrt{|\mathcal{N}(i)|}\sqrt{|\mathcal{N}(j)|}$. $\sigma$ represents a nonlinear activation function. In addition to GCN, commonly used GNN models include GraphSAGE[35] and GAT[36]. These models differ from GCN mainly in their information aggregation and feature updating methods, which will be further introduced in the Method section.

## 2.3 Solving combinatorial optimization problems with GNNs

In order to enable GNNs to solve COPs in an unsupervised manner, a recently proposed method, PI-GNN[12], transforms the Hamiltonian $H$ of a COP into a differentiable loss function $loss_H(\theta)$. The key to this transformation is to replace the decision variable $x_i$ by $p_i(\theta)$, where $\theta$ represents all parameters in a GNN and $p_i(\theta) \in [0,1]$ represents the final output of the GNN model on node $i$. Based on this transformation, the loss function of the network model can be defined as

$$loss_H(\theta) = \sum_{i,j} p_i(\theta) Q_{ij} p_j(\theta). \tag{4}$$

Specifically, one can input randomly initialized embedding node features $h_i^{(0)}$ and train a GNN with the CGBP algorithm to obtain the final output $p_i(\theta)$. Then, a projection strategy can be used to map $p_i(\theta)$ to the decision variable $x_i$, which generates the solution of the COP. The simplest projection strategy is to use a threshold of 0.5, where $p_i(\theta)$ above the threshold are set to 1 and values below the threshold are set to 0.

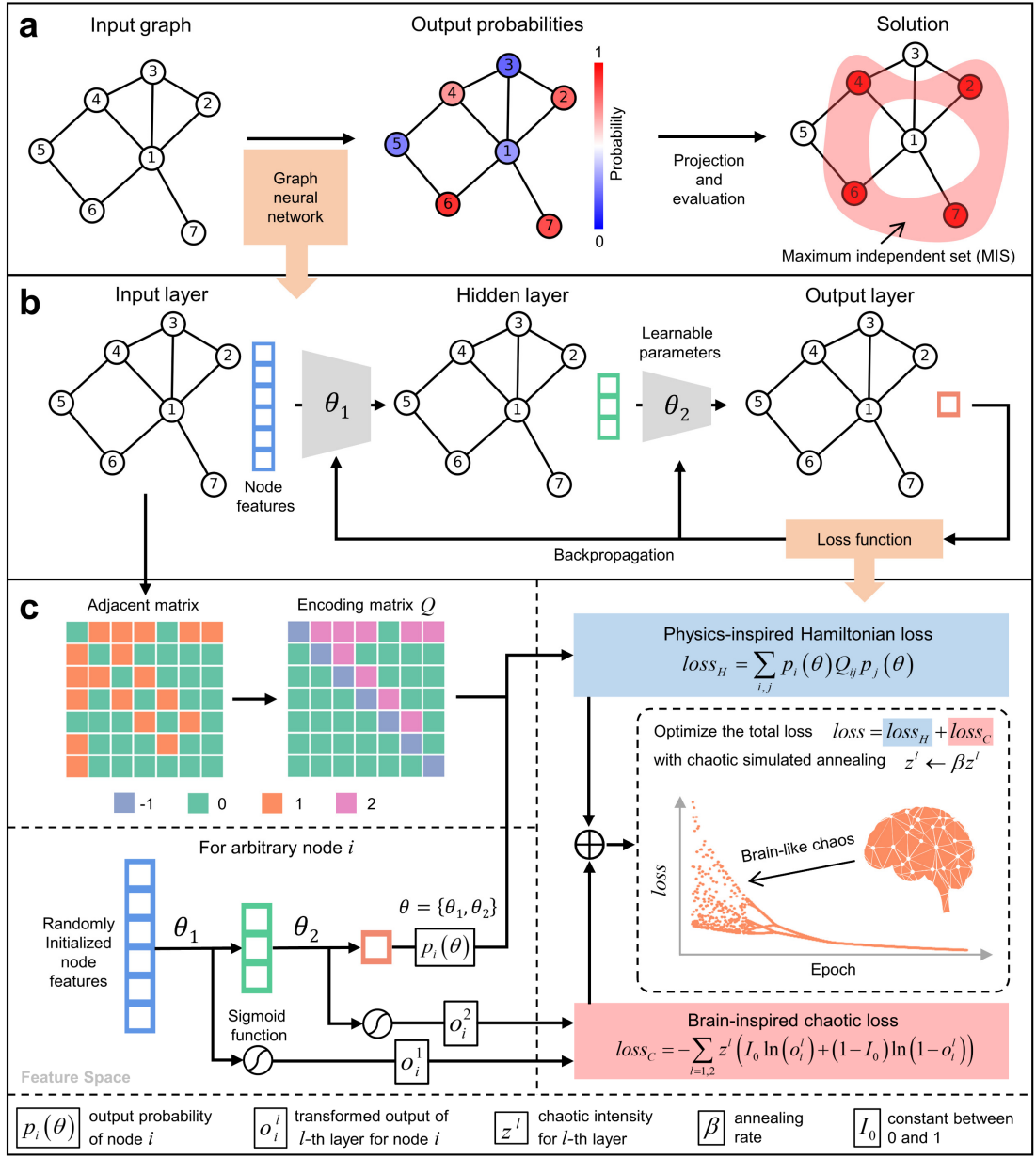

**Fig. 1. Schematic diagram of CGBP. a** shows the procedure of using GNN to solve the MIS problem. First, solving the MIS problem is mapped to the training process of GNN, and the probability of each node belonging to the MIS is obtained after training. Finally, the probability values are projected to the solution of the MIS problem by setting a proper threshold. **b** shows a simple two-layer GNN with trainable parameters denoted by $\theta = \{\theta_1, \theta_2\}$. **c** shows that the loss function of GNN consists of two parts: the physics-inspired Hamiltonian loss $loss_H$ that encodes the MIS problem, and the brain-inspired chaotic loss $loss_C$ introduced by CGBP.

## 2.4 Chaotic graph backpropagation

Recently, we proposed a chaotic backpropagation (CBP)[37] algorithm for multilayer perceptron (MLP), which introduces a loss function to simulate chaotic dynamics in the brain. Here, we use a similar strategy as CBP to construct the chaotic graph backpropagation (CGBP) algorithm. That is, an additional chaotic loss function $loss_C$ is added to the original loss function $loss_H$, given by

$$loss_C = -\sum_l \sum_j z_j^{(l)} \left( I_0 \ln \left( o_{dj}^{(l)} \right) + (1 - I_0) \ln \left( 1 - o_{dj}^{(l)} \right) \right), \tag{5}$$

where $z_j^{(l)}$ is the corresponding chaotic strength for $o_{dj}^{(l)}$, $I_0$ is a constant between 0 and 1 (usually set to 0.65), $o_{dj}^{(l)}$ is the $j$th element of $o_d^{(l)}$, the intermediate output of node $d$ in the $l$th layer. The mathematical expression of $o_d^{(l)}$ for the GCN in Eq. (3) is given as follows:

$$o_d^{(l)} = sigmoid \left( \sum_{k \in \mathcal{N}(d)} \frac{1}{c_{dk}} h_d^{(l-1)} W^{(l)} + B^{(l)} \right). \tag{6}$$

It should be noted that compared to the $loss_C$ introduced in CBP, the $loss_C$ in CGBP needs to consider not only the output of the neurons but also the selection of node $d$, because in GNN all nodes share the same parameters $\theta$, which makes CGBP more complex than CBP. To further explain the mechanism that how $loss_C$ generates chaotic dynamics, we calculate the gradient of $loss_C$ with respect to the weight $w_{ij}^{(l)}$,

$$\begin{aligned}
\frac{\partial loss_C}{\partial w_{ij}^{(l)}} &= -z_j^{(l)} \left( \frac{I_0}{o_{dj}^{(l)}} \frac{\partial o_{dj}^{(l)}}{\partial w_{ij}^{(l)}} - \frac{(1-I_0)}{(1-o_{dj}^{(l)})} \frac{\partial o_{dj}^{(l)}}{\partial w_{ij}^{(l)}} \right) \\
&= -z_j^{(l)} \frac{I_0 - o_{dj}^{(l)}}{o_{dj}^{(l)} (1 - o_{dj}^{(l)})} o_{dj}^{(l)} (1 - o_{dj}^{(l)}) \frac{1}{c_{dk}} \sum_k h_{ki}^{(l-1)} \\
&= -\frac{1}{c_{dk}} \sum_k h_{ki}^{(l-1)} z_j^{(l)} \left( I_0 - o_{dj}^{(l)} \right).
\end{aligned} \tag{7}$$

Similarly, the gradient of the bias $\partial loss_C / \partial b_i^{(l)}$ can also be obtained, but for simplicity, we will not discuss it here. At this point, after introducing $loss_C$, the total loss of the model is $loss = loss_H + loss_C$, and in the standard BP algorithm, the updating dynamics of $w_{ij}^{(l)}$ can be expressed as

$$\begin{aligned}
w_{ij}^{(l)} &\leftarrow w_{ij}^{(l)} - \eta \frac{\partial (loss_H + loss_C)}{\partial w_{ij}^{(l)}} \\
&= w_{ij}^{(l)} - \eta \frac{\partial loss_H}{\partial w_{ij}^{(l)}} - \eta \frac{\partial loss_C}{\partial w_{ij}^{(l)}} \\
&= w_{ij}^{(l)} - \eta \frac{\partial loss_H}{\partial w_{ij}^{(l)}} + \frac{\eta}{c_{dk}} \sum_k h_{ki}^{(l-1)} z_j^{(l)} \left( I_0 - o_{dj}^{(l)} \right),
\end{aligned} \tag{8}$$

where $\eta$ is the learning rate. Considering that $z_j^{(l)}$ is a tunable parameter, we can always define a new $z_j^{(l)}$, so that $z_j^{(l)} \leftarrow \frac{\eta}{c_{dk}} \sum_k h_{ki}^{(l-1)} z_j^{(l)}$, and then Eq. (8) can be simplified as

$$w_{ij}^{(l)} \leftarrow w_{ij}^{(l)} - \eta \frac{\partial loss_H}{\partial w_{ij}^{(l)}} + z_j^{(l)} \left( I_0 - o_{dj}^{(l)} \right). \tag{9}$$

From a dynamical viewpoint, we can describe these updating equations (9) as difference equations in a vector or matrix form when taking iteration number as time $t$, i.e.

$$W(t+1) = F(W(t)) \tag{10}$$

where $W(t)$ is the vector representing all $w_{ij}^{(l)}$ at the iteration $t$, and $F$ is the corresponding function vector describing the right-hand side of Eq. (9). Thus, we can analyze this updating process from the perspective of dynamical systems. From the above equation, it can be seen that the introduction of $loss_C$ is equivalent to adding negative feedback to the weight $w_{ij}^{(l)}$ (note that $o_{dj}^{(l)}$ is the monotone function of $w_{ij}^{(l)}$), and it is theoretically proven that when $z_j^{(l)}$ is sufficiently large, the dynamics in Eq. (9) exhibit Marotto chaos (high-dimensional topological chaos)[31]. It should be noted that the dynamics in Eq. (9) is based on the Nagumo-Sato[38] model and the chaotic neuron model[39], which models real neurons, thus the chaotic term or chaotic dynamics in Eq. (9) has biological significance. In addition, although Eq. (9) is derived from the standard BP algorithm, it can be easily extended to other improved versions, such as SGDM[40] and Adam[13], as detailed in Supplementary Information.

On the other hand, in order to ensure the convergence of the learning process, we adopt a chaotic simulated annealing strategy[18], that is

$$z_j^{(l)} \leftarrow \beta z_j^{(l)}, \tag{11}$$

where $\beta$ is an annealing constant close to 1 (such as 0.999), and in this work, we set all $z_j^{(l)}$ to the same value $z$. Similarly, we can also describe Eq. (11) as difference equations $z(t+1) = \beta z(t)$, and analyze their dynamical features combined with the above difference equations $W(t+1) = F(W(t))$.

Clearly, the proposed updating dynamics (9) and (11) begin with chaotic dynamics due to high $z$, then enter a bifurcation process with intermediate $z$ with the evolution of iteration $t$, and eventually reduce to gradient dynamics (or traditional BP updating dynamics) when $z$ is sufficiently small. As shown in the additional term of Eq. (9), another significant feature is that CGBP can be added to any existing algorithm as a plug-in unit for improving its performance, thus it can be considered as a universal learning algorithm for GNNs.

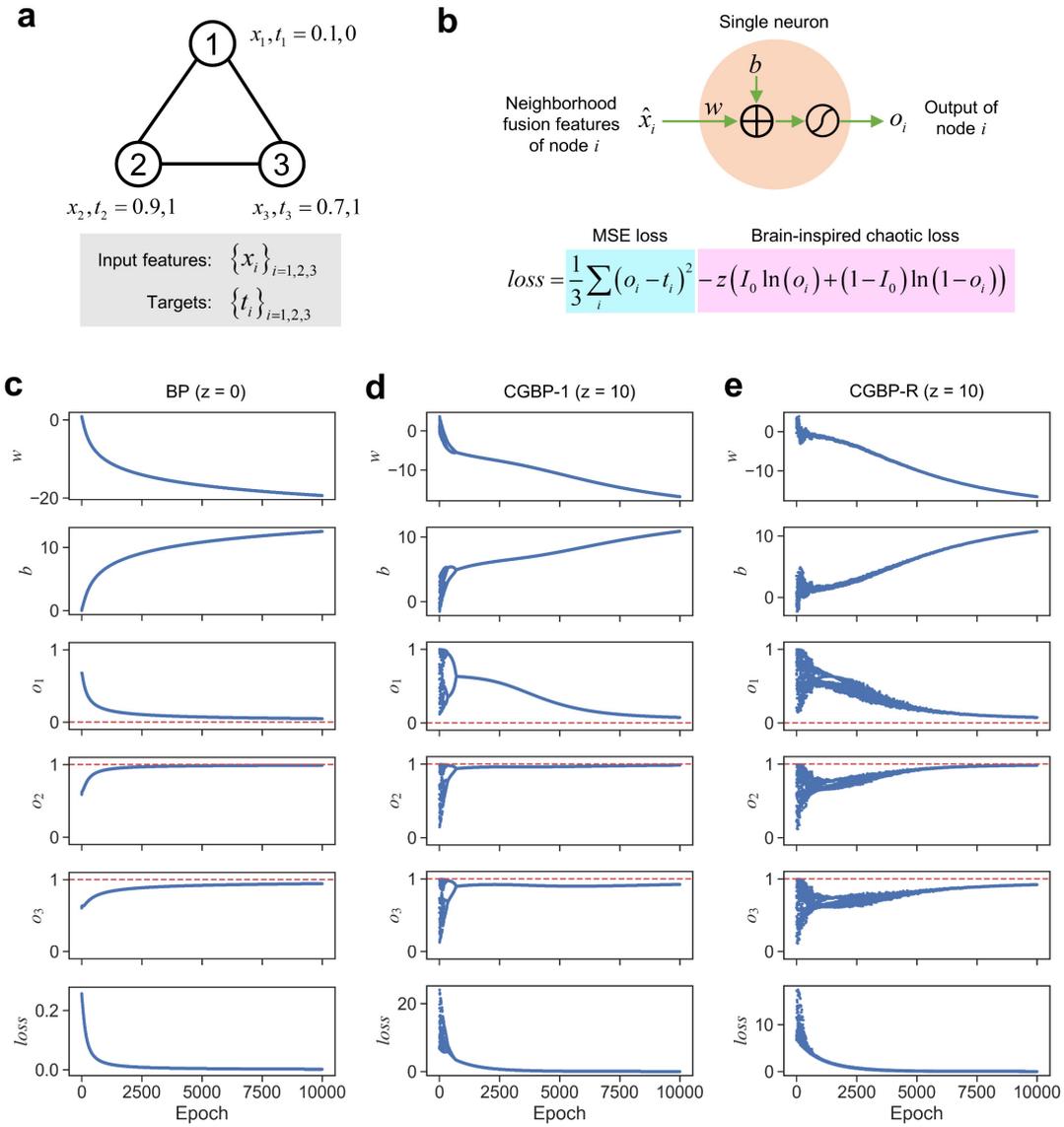

Fig 2. Numerical results on 2-regular graphs. a shows an example of a 3-node 2-regular graph, where the input features and targets (labels) of the nodes are labeled. b shows a single-neuron model used to train the example graph in a, where the total loss function of the model is defined as the sum of a mean squared error (MSE) loss (blue) and a chaotic loss (pink). c shows the changes in weights ($w$ and $b$), the network output ($o_{1\sim3}$) and total loss (*loss*) when the initial chaotic strength $z$ is 0 (equivalent to BP). d and e show when $z$ is set to 10, the training processes of CGBP-1 and CGBP-R, respectively, with the former having $o_i = o_1$ in the chaotic loss during training and the latter randomly selecting from $o_{1\sim3}$.

# 3   Results

## 3.1   Testing the performance of CGBP on a 3-node graph

In order to illustrate the working principle of the CGBP algorithm, we take a 2-regular graph with 3 nodes shown in Fig. 2a as an example to verify the chaotic dynamics and convergence of CGBP. As an example, the input features of these three nodes are set to 0.1, 0.9 and 0.7, respectively, and their corresponding target values (labels) are set to 0, 1 and 1, respectively. For any node $i$, if we use the single neuron model shown in Fig. 2b, we can define a loss function that measures the difference between the output of node $i$ and the target, such as the mean squared error (MSE) loss, to learn the model parameters $w$ and $b$. Based on this, the CGBP algorithm introduces a brain-inspired chaotic loss, and we first verify that this chaotic loss can cause the parameters $w$ and $b$ training process to exhibit chaotic dynamics. As shown in Fig. 2c, when the chaotic strength $z = 0$, the CGBP algorithm is equivalent to the traditional BP algorithm. At this time, both $w$ and $b$ are updated using continuous gradient dynamics, and the node output $o_i$ and total loss (*loss*) are also updated in the same way. When $z = 10$ ($i$ is always set as 1, denoted as CGBP-1) i.e. high $z$, in the early training stages (epoch < 1000), the dynamics of $w$ and $b$ is discrete (as shown in Fig. 2d). Numerically, it can be found that the dynamics has positive Lyapunov exponents, which means their dynamics is chaotic (as shown in Supplementary Fig. S1). In addition, from Fig. 2d, it can be seen that as $z$ is annealed (with annealing constant $\beta = 0.999$), the chaotic dynamics of $w$ and $b$ gradually disappears through a series of bifurcations (with middle $z$) and degenerate into gradient dynamics (with small $z$), and finally converge as $z$ approaches 0. It should be noted that in each updating of $w$ and $b$, $i$ can be set as a constant value or randomly selected from $\{1, 2, 3\}$ (as shown in Fig. 2e, denoted as CGBP-R). At this time, in addition to gradient and chaotic dynamics, $w$ and $b$ are also affected by stochastic dynamics, but they still converge with the annealing of *z*. Considering that these two selected ways of *i* have little impact on the performance of CGBP in practical use, we do not distinguish between them later.

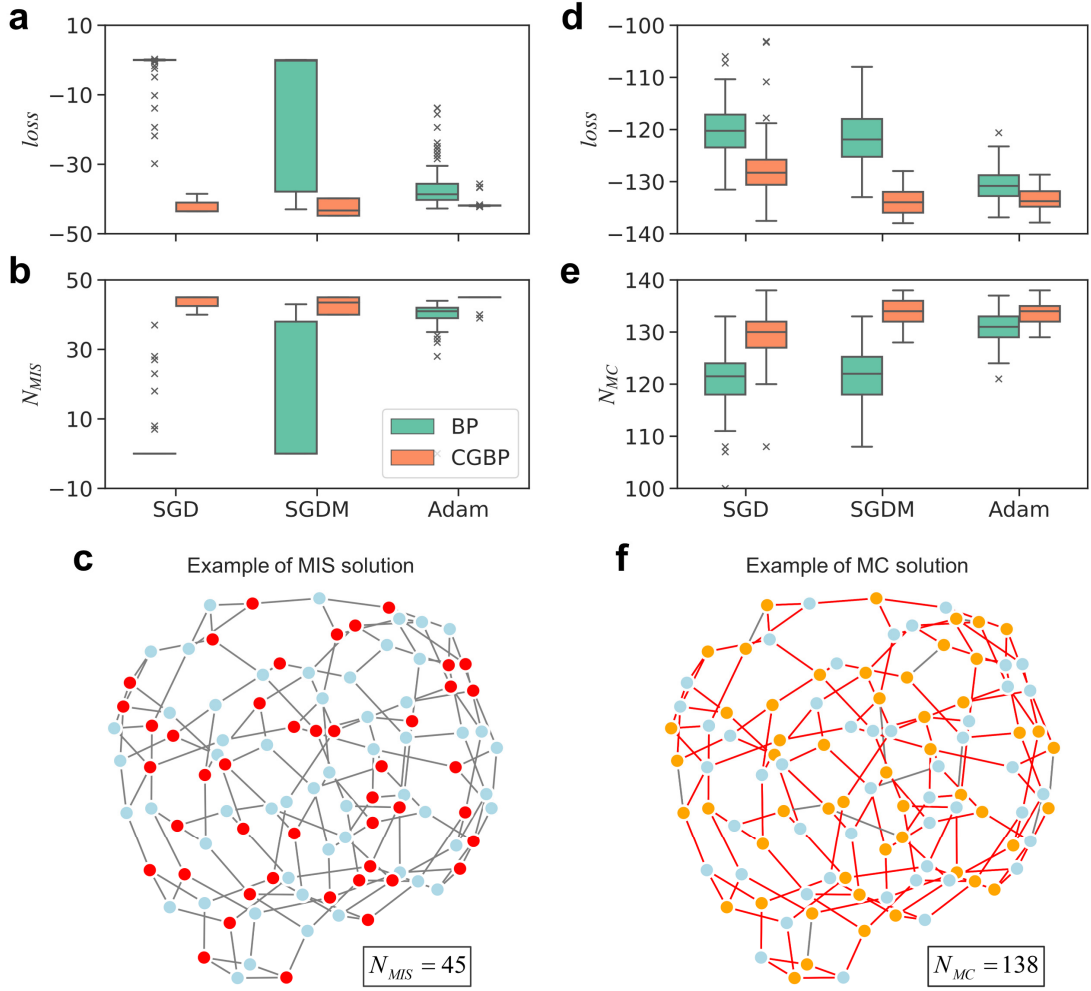

**Fig 3. Numerical results on a 3-regular graph with 100 nodes.** For the MIS problem, **a** and **b** compare the minimum loss (*loss*) and the size of the MIS ($N_{MIS}$) achieved by BP and CGBP methods when using SGD, SGDM and Adam optimizers. **c** shows a solution obtained by CGBP for the MIS problem (red nodes). For the MC problem, **d** and **e** compare *loss* and the maximum cut size ($N_{MC}$) achieved by BP and CGBP methods when using different optimizers. **f** shows a solution obtained by CGBP for the MC problem (red edges). Each scenario in the boxplots represents the results of 100 independent training runs, where '×' indicates outliers.

## 3.2 Testing the performance of CGBP on large-scale 3-regular graphs

In the previous section, we have verified the chaotic dynamics and convergence of CGBP on a toy graph. Next, we further test the performance and robustness of CGBP on large-scale 3-regular graphs.

First, we discuss the MIS problem on a 3-regular graph with 100 nodes. We selected three BP-based optimization algorithms, SGD, SGDM, and Adam. For each optimization algorithm, we trained a GCN model (see Methods for details) 100 times with different random number seeds, and recorded the minimum *loss* and the maximum MIS (where the size of the set is denoted as $N_{MIS}$) that the algorithm can reach during the training process. In addition, by introducing the chaotic loss, we implemented the CGBP versions of these three optimization algorithms and performed similar statistics. In Figs. 3a and 3b, the minimum *loss* and $N_{MIS}$ of BP and CGBP methods are compared under different optimizers. It can be seen that regardless of the optimizer used, the minimum *loss* that CGBP can achieve is significantly smaller than that of BP, and $N_{MIS}$ is greater than that of BP. Moreover, the performance of BP on these two indicators significantly depends on the optimization algorithm, while CGBP has strong and stable performance under different optimization algorithms. Fig. 3c shows the solution obtained by the CGBP algorithm for an MIS problem, where the red point set forms the final MIS with corresponding $N_{MIS} = 45$. In addition to the MIS problem, we also discussed the MC problem (see Methods for the detailed definition) on this graph. Figs. 3d and 3e compare the minimum *loss* and maximum cut ($N_{MC}$) of the BP and CGBP methods under different optimizers for the MC problem, and it can be seen that the result obtained in the MC problem is highly consistent with that in the MIS problem. Fig. 3f shows the solution obtained by the CGBP algorithm for an MC problem, which is the set of red edges, with $N_{MC} = 138$.

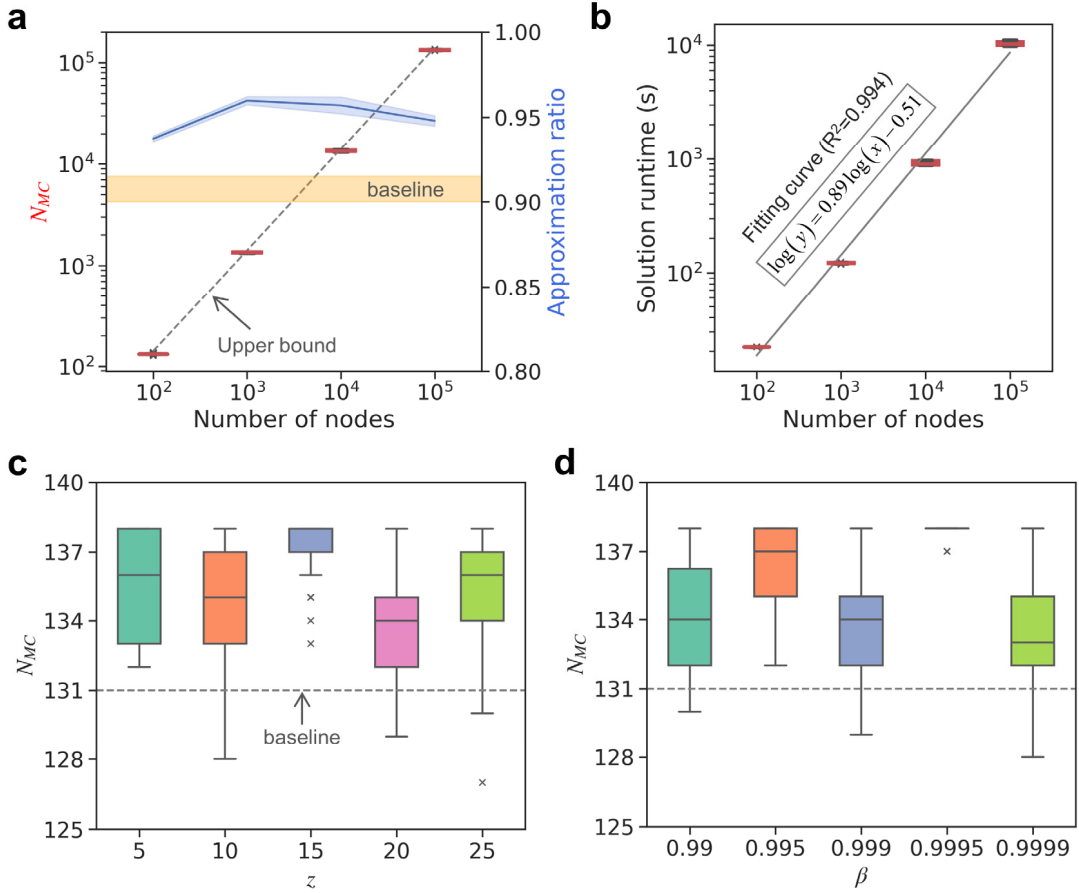

**Fig. 4. Numerical results on large-scale 3-regular graphs for MC problem. a** shows the statistical results of the $N_{MC}$ obtained by CGBP in 100 independent training runs as the number of nodes $n$ increases from $10^2$ to $10^5$, where the approximation ratio of $N_{MC}$ to the theoretical upper bound (gray dashed line) is illustrated by the blue line. The orange area represents the range of the approximation ratios for the baseline (the original BP-based PI-GNN). **b** shows the statistical results of the running time required for CGBP to solve the corresponding MC problem as the number of nodes increases (hardware is described in the Method section). A clear linear relationship (gray line) in this region can be fitted by taking the logarithm of the number of nodes as $x$ and the average running time as $y$. **c** and **d** show the statistical results of $N_{MC}$ obtained by CGBP with different $z$ (fixed $\beta$) in a 3-regular graph with 100 nodes and with different $\beta$ (fixed $z$), where the baseline is the median of $N_{MC}$ obtained by training with BP (gray dashed line).

Next, we analyze in depth the performance of the CGBP method on larger 3-regular graphs using the MC problem as an example. Fig. 4a shows the change of $N_{MC}$ as the number of nodes $n$ increases. It can be theoretically proven[41] that for a $d$-regular graph with $n$ nodes, the upper bound of $N_{MC}$ can be approximated by

$Cut_{ub} = \left( d/4 + P^* \sqrt{d/4} \right) n$, where $P^* \approx 0.7632$ is a constant related to the ground state energy of the Sherrington-Kirkpatrick model[42]. To evaluate the performance of the CGBP method, we define an approximation ratio, that is $N_{MC}/Cut_{ub}$, where a value closer to 1 indicates a higher quality of the solution. As can be seen, as $n$ increases from $10^2$ to $10^5$, the approximation ratio of CGBP remains around 0.95 throughout, significantly higher than the original PI-GNN based on BP, which ranges from 0.9 to 0.92 (orange area). This indicates that the chaotic dynamics introduced by CGBP can significantly improve the quality of the solution for large-scale MC problems. It is worth noting that since CGBP only introduces one loss function, compared to the entire training process, it does not bring significant computational burden, so the training time of PI-GNN based on CGBP is not significantly different from that based on BP (as shown in Supplementary Fig. S2). In particular, the running time of CGBP increases linearly with the number of nodes $n$ (as shown in Fig. 4b), that is, the time complexity of the algorithm is $O(\sim n)$, significantly lower than traditional methods such as the Goemans-Williamson[43] algorithm based on semi-definite programming with a time complexity of $O(\sim n^{3.5})$.

On the other hand, although the chaotic loss introduces two hyperparameters $z$ and $β$ in CGBP, the improvement in optimization performance is robust to both of them. Taking the example of the MC problem on a 3-regular graph with 100 nodes, we first fixed $β = 0.999$ and then counted the $N_{MC}$ of CGBP in 100 independent training runs with initial chaotic strength $z$ set to 5, 10, 15, 20, and 25 (Fig. 4c), with corresponding median $N_{MC}$ values of 136, 135, 138, 134 and 136, respectively. Similarly, we fixed $z = 20$ and then counted the $N_{MC}$ of CGBP in 100 independent training runs with $β$ set to 0.99, 0.995, 0.999, 0.9995 and 0.9999 (Fig. 4d), with corresponding median $N_{MC}$ values of 134, 137, 134, 138 and 133, respectively. It can be seen that, regardless of various hyperparameter combinations, CGBP has a significant and consistent improvement compared to the median $N_{MC}$ of 131 in BP (gray dashed line).

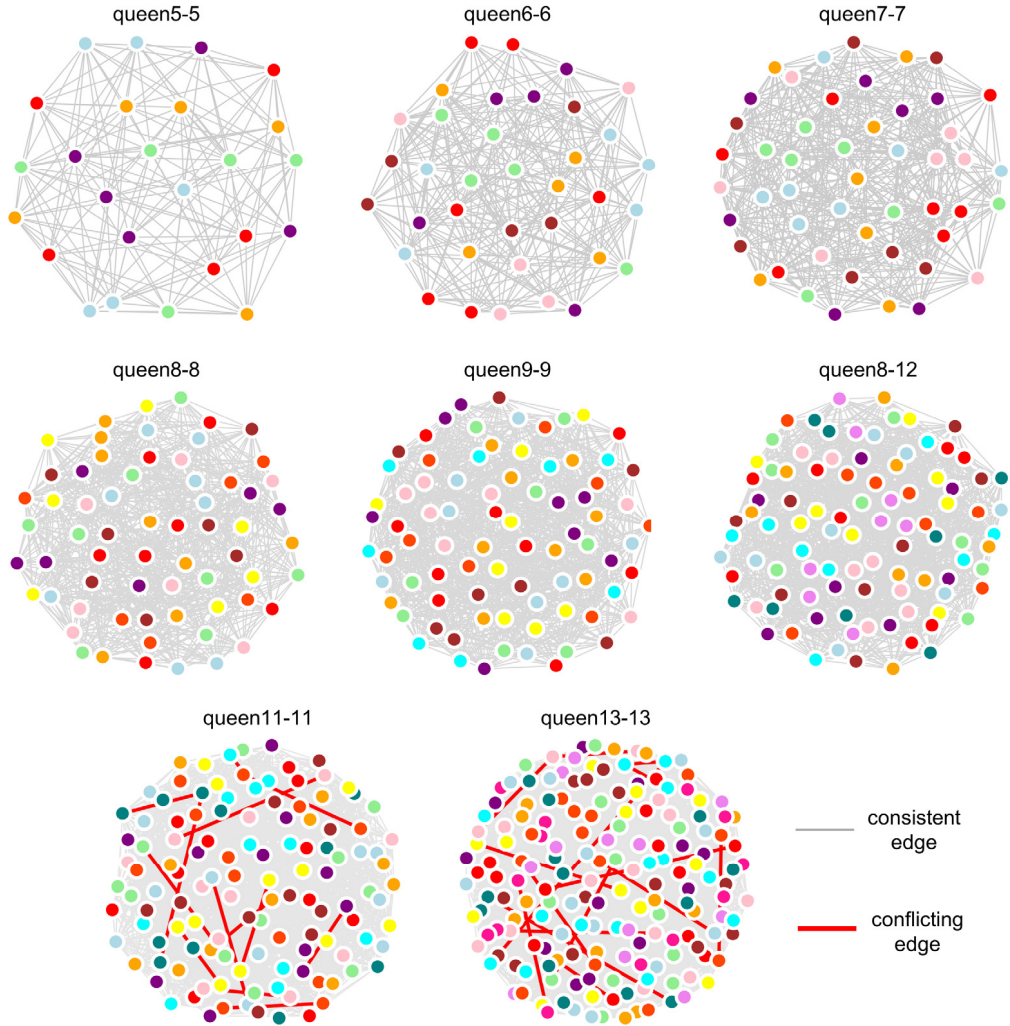

**Fig. 5. Graph coloring results of CGBP on the Queen dataset.** The GNN architecture used here is GraphSAGE. The edges with different or the same colors on both vertexes (consistent or inconsistent edges) are represented by gray and red solid lines, respectively.

### 3.3 Comparison with existing SOTA methods

To further test the performance of CGBP on the MC task, we compared it with existing SOTA methods on 5 instances in the Gset dataset, including BLS[44], DSDP[45], KHLWG[46], RUN-CSP[47], and PI-GNN[12, 48] (which includes two versions, PI-GCN and PI-SAGE, corresponding to the GCN and GraphSAGE architectures, respectively).

Table 1 shows the best $N_{MC}$ obtained by different methods on these instances (the training processes are shown in Supplementary Figs. S3 and S4). It can be seen that after training with CGBP, both PI-GCN and PI-SAGE have significant improvements in $N_{MC}$ overall. Furthermore, compared to existing SOTA methods, such as BLS and KHLWG based on local search and taboo search respectively, the difference between CGBP and them in terms of $N_{MC}$ is only 29~41 on instances G14, G15, and G22. Notably, CGBP also achieved the optimal $N_{MC}$ on the larger instances G49 and G50. These results indicate that CGBP has a performance that is not only better than existing deep learning methods but also comparable to the existing SOTA methods on large scale MC problems, thus further demonstrating the effectiveness of CGBP on the MC task.

**Table 1.** Comparison of $N_{MC}$ of different algorithms on the Gset dataset. The best results are bolded.

| Graph | Nodes | Edges | BLS | DSDP | KHLWG | RUN-CSP | PI-GCN | PI-SAGE | PI-GCN +CGBP | PI-SAGE +CGBP |
|---|---|---|---|---|---|---|---|---|---|---|
| G14 | 800 | 4694 | **3064** | 2922 | 3061 | 2943 | 3026 | 3031 | 3035 | 3029 |
| G15 | 800 | 4661 | **3050** | 2938 | **3050** | 2928 | 2990 | 3006 | 3003 | 3016 |
| G22 | 2000 | 19990 | **13359** | 12960 | **13359** | 13028 | 13181 | 13125 | 13177 | 13318 |
| G49 | 3000 | 6000 | **6000** | **6000** | **6000** | **6000** | 5918 | 5894 | **6000** | **6000** |
| G50 | 3000 | 5880 | **5880** | **5880** | **5880** | **5880** | 5820 | 5844 | 5878 | **5880** |

**Table 2.** Comparison of $N_{GC}$ of different algorithms on the Queen dataset. The best results are bolded.

| Graph | Nodes | Edges | Colors | Tabucol | Tabucol2 | GNN | PI-GCN | PI-SAGE | PI-GCN +CGBP | PI-SAGE +CGBP |
|---|---|---|---|---|---|---|---|---|---|---|
| queen5-5 | 25 | 160 | 5 | **0** | **0** | **0** | **0** | **0** | **0** | **0** |
| queen6-6 | 36 | 290 | 7 | **0** | **0** | 4 | 1 | **0** | 1 | **0** |
| queen7-7 | 49 | 476 | 7 | **0** | 10 | 15 | 8 | **0** | 5 | **0** |
| queen8-8 | 64 | 728 | 9 | **0** | 8 | 7 | 6 | 1 | 3 | **0** |
| queen9-9 | 81 | 1056 | 10 | **0** | 5 | 13 | 13 | 1 | 3 | **0** |
| queen8-12 | 96 | 1368 | 12 | **0** | 10 | 7 | 10 | **0** | 4 | **0** |
| queen11-11 | 121 | 1980 | 11 | 20 | 33 | 33 | 37 | 17 | 20 | **13** |
| queen13-13 | 169 | 3328 | 13 | 35 | 42 | 40 | 61 | 26 | 31 | **15** |

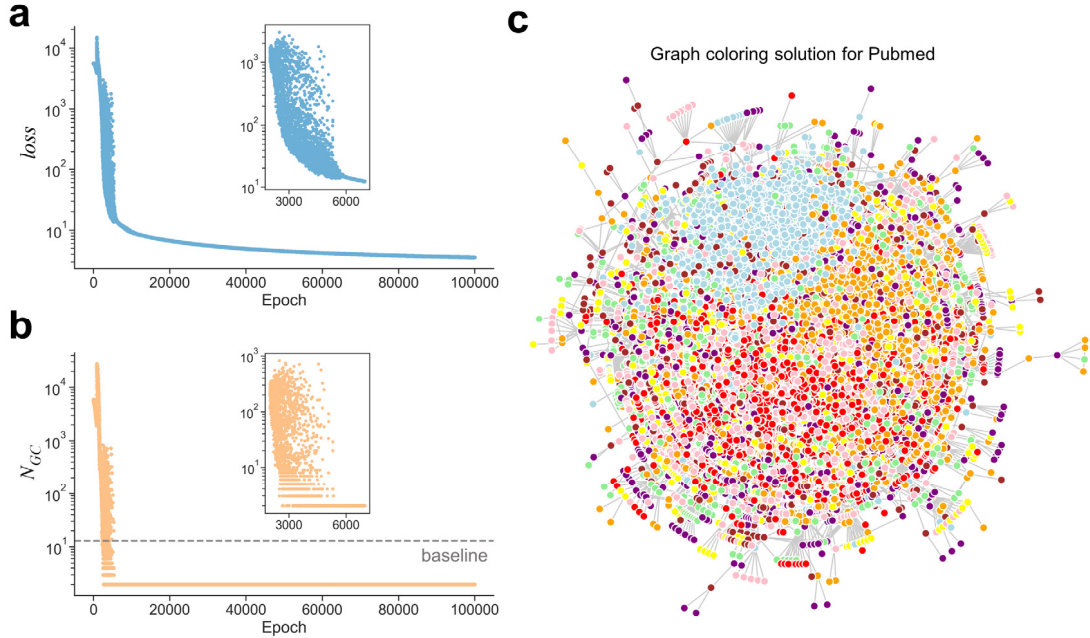

**Fig. 6. Graph coloring results of CGBP on the Pubmed graph. a** and **b** show the changes in *loss* and $N_{GC}$ during the process of training PI-SAGE using CGBP. The gray dashed line refers to the minimum $N_{GC}$ obtained by BP-based PI-GNN. **c** shows the coloring result of Pubmed corresponding to $N_{GC} = 2$.

**Table 3.** Comparison of $N_{GC}$ of different algorithms on the Citation dataset, with the best results highlighted in bold. '-' means the method failed to obtain results within 24 hours.

| Graph | Nodes | Edges | Colors | Tabucol | Tabucol2 | GNN | PI-GCN | PI-SAGE | PI-GCN +CGBP | PI-SAGE +CGBP |
|---|---|---|---|---|---|---|---|---|---|---|
| Cora | 2708 | 5429 | 5 | **0** | 31 | 3 | 1 | **0** | **0** | **0** |
| Citeseer | 3327 | 4732 | 6 | **0** | 6 | 3 | 1 | **0** | 1 | **0** |
| Pubmed | 19717 | 44338 | 8 | - | - | 35 | 13 | 17 | 9 | **2** |

The previous MIS and MC tasks are equivalent to the binary classification of nodes in the graph. In fact, CGBP can also be applied to more complex multi-classification tasks. Here, we take the GC problem as an example, whose task objective is to color the nodes in the graph with a given number of colors, so that the number of edges ($N_{GC}$) whose both vertexes have the same color is minimized. The dataset used for testing includes two public datasets, one is the Queen dataset with 25~169 nodes but high network density (23.44%~54.44%), and the other is the Citation dataset with 2708~19717 nodes but low network density (0.02%~0.15%, including three graph instances: Cora[49], Citeseer[50] and Pubmed[51]). Table 2 summarizes the results of CGBP and existing SOTA methods (including Tabucol[52], Tabucol2[53], and GNN[53]) on the Queen dataset. It can be seen that after training with CGBP, the $N_{GC}$ of both PI-GCN and PI-SAGE decreases to different degrees, especially on the largest two instances queen11-11 and queen13-13. In addition, PI-SAGE trained with CGBP achieves the best $N_{GC}$ on different instances. Fig. 5 shows the coloring results of PI-SAGE+CGBP on the Queen dataset (the training processes are shown in Supplementary Figs. S5 and S6). Similar to the results on the Queen dataset, CGBP also achieves the best performance on the three citation graph instances, Cora, Citeseer and Pubmed (Table 3, the corresponding training processes and coloring results can be found in Supplementary Figs. S7, S8 and Fig. 6), especially the combination of PI-SAGE and CGBP achieves $N_{GC}$ = 2 on the Pubmed instance, reducing the number of errors by an order of magnitude compared to existing methods (Fig. 6). The above results indicate that the CGBP method has excellent performance in solving large-scale GC problems, topping the SOTA methods.

# 4  Discussion

In this work, we draw inspiration from the chaotic dynamics in the real brain and develop a universal learning algorithm for GNNs, which is applied to PI-GNN to efficiently solve large-scale COPs, outperforming the SOTA methods. Compared with existing combinatorial optimization methods, the CGBP method has the following advantages: (1) The optimization performance of CGBP on multiple publicly available benchmark datasets can surpass existing SOTA methods; (2) The time complexity of CGBP is linear, that is, the computation time increases linearly with the increase of the system variables, enabling it to handle super-large-scale COPs

(currently it can handle graphs of millions of nodes[12], but note that the obtained solution may not optimal); (3) CGBP is robust, that is, its optimization performance is less affected by the hyperparameters $z$ and $\beta$; (4) Compared with other machine learning-based methods, CGBP directly transforms the solving process of a COP into the GNN training process, hence there is no need to give the training set in advance; (5) CGBP can be directly extended to other COPs, such as the constraint satisfaction problem (CSP); (6) CGBP, as a universal learning algorithm for GNNs, can be used as a plug-in unit to easily integrate into any existing algorithm for improving global searching ability and performance.

Specifically, we believe that the strong optimization performance of CGBP can be attributed to three main reasons. The first reason is that chaotic dynamics has been widely used to solve optimization problems due to its theoretically guaranteed global ergodicity and pseudo-randomness[15-18], and existing theory has shown that the chaotic loss introduced by the CGBP method can induce chaotic dynamics when $z$ is sufficiently large[19], which generates Marotto chaos and thus makes the training dynamics rich and global. The second reason is that the chaotic loss introduced by CGBP is inspired by the real brain, which is more biologically plausible compared to other methods that generate chaos by adjusting the learning rate or introducing chaotic mappings[54]. It should be noted that the chaotic loss introduced by CGBP is a local loss, which is equivalent to introducing local negative feedback to the weight parameters. This local regulatory feature is consistent with some biological experiments on synaptic plasticity mechanisms, such as spike-timing-dependent plasticity (STDP)[55]. Interestingly, the recently proposed Forward-Forward learning algorithm by Hinton also uses local losses[56], and the use of local learning may be more competitive in the future. The third reason is that the chaotic loss introduced by CGBP has the form of cross-entropy, so when the chaotic loss dominates in the early stages of learning, the output of neurons tends to be $I_0$, which may make the network more plastic. Furthermore, from an optimization perspective, this is similar to the idea of interior point methods for solving linear programming problems.

Although CGBP has made important progress in COPs, there is still a lot of room for improvement in its optimization performance. Here are some possible ways to further improve it. Firstly, the hyperparameters $z$ and $\beta$ in the CGBP method need to be set manually, and how to adaptively adjust these two hyperparameters to achieve the optimal performance of CGBP is still an unresolved issue. Secondly, the optimization algorithm and architecture of the neural network also have a certain impact on the optimization performance of CGBP. In this work, we only tried SGD, SGDM and Adam optimizers, and considered only GCN and GraphSAGE architectures. So far, the combination of Adam and GraphSAGE has achieved the best results. Using more advanced optimizers, such as Adablief[57] and Adan[58], and GNN architectures, such as GAT[36] and SGAT[59], may further improve the performance of CGBP. Finally, the physics-inspired Hamilton loss used in PI-GNN may not correspond well to the solutions of COPs, and sometimes the solution with a lower loss has a worse quality. Therefore, how to encode the solution of combinatorial optimization more rationally is also an important research direction in the future.

# 5  Methods

## 5.1  Encoding of MC and GC problems

The objective of the MC problem is to divide the vertex set $\mathcal{V}$ of a given graph $\mathcal{G}(\mathcal{V},\mathcal{E})$ into two subsets $\mathcal{V}_1$ and $\mathcal{V}_2$, where the number of edges ($N_{MC}$) between the vertex subset $\mathcal{V}_1$ and its complementary subset $\mathcal{V}_2$ is maximized. Mathematically, the corresponding Hamiltonian can be written as[12]

$$H_{MC} = \sum_{i<j} A_{ij}\left(2x_i x_j - x_i - x_j\right). \tag{12}$$

The GC problem is an extension of the MC problem, where the dimension of the variables is expanded to multiple dimensions ( $x_i \in \{1, 2, \cdots, q\}$ ). Simply put, the objective of the GC problem is to assign one of $q$ colors to each node, such that the number of edges ($N_{GC}$) with the same color on both vertexes is minimized. The corresponding Hamiltonian of the Potts model is[48]

$$H_{GC} = -J \sum_{(i,j) \in \mathcal{E}} \delta(x_i, x_j), \tag{13}$$

where $J$ represents the strength of interaction between nodes, which can be directly set to -1 here, and $\delta(x_i, x_j)$ is the Kronecker delta, which equals 1 when $x_i = x_j$ and 0 otherwise. It should be noted that for the GC problem, the activation function of the last layer of GNN needs to be replaced by softmax from sigmoid. Therefore, $p_i(\theta)$ is no longer a scalar from 0 to 1, but a $q$-dimensional vector, where different dimensions represent the probability of assigning different colors.

## 5.2  Architecture of GNNs

This work used two classic GNN architectures: GCN and GraphSAGE. The architecture was built based on the DGL package[60]. Compared to the transductive learning of GCN, GraphSAGE is an inductive learning framework that efficiently uses node attribute information to generate embeddings for new nodes. The calculation and updating process can be written as[35]

$$h_i^{(l+1)} = \sigma\left(W^{(l)} \cdot concat\left(h_i^{(l)}, aggregate\left(\{h_j^{(l)}, \forall j \in \mathcal{N}(i)\}\right)\right)\right), \tag{14}$$

where $concat(\cdot)$ represents the concatenation of feature vectors, and $aggregate(\cdot)$ represents the aggregation of neighbor feature vectors. In this paper, we applied the average aggregation. In addition, the number of GNN layers was two, where the input and output dimensions of the first layer were $d_0$ and $d_1$, respectively, and the second layer were $d_1$ and $d_2$, respectively. These dimensions were set using a similar scheme to PI-GNN[12]. Specifically, for a graph with $n$ nodes, we set $d_0 = int(\sqrt{n})$ and

$d_1 = \text{int}(d_0/2)$. For binary classification tasks such as MIS and MC, we set $d_2 = 1$, and for multi-classification tasks such as GC, we set $d_2 = q$. We also applied activation functions between the layers, where the first layer was ReLU[61] and the second layer was sigmoid (softmax for multi-classification). To further improve GNN performance, we added Dropout[62] and BatchNorm[63] in each layer.

### 5.3  Setting of training parameters

Since the training results depend on the initialization of the weight parameters, when comparing with other methods, we used different random seeds for initialization, then trained multiple times and took the best result. The model weights were initialized using the default initialization method in Pytorch[64], where the weight of layer $i$ is randomly initialized from a uniform distribution in the interval $\left[-\sqrt{k_i}, \sqrt{k_i}\right]$, where $k_i = 1/M_{i-1}$ is the reciprocal of the number of input neurons in the layer. For 3-regular graphs of different sizes, the maximum training epochs were set to 10000, while for Gset, Queen, and Citation datasets, the maximum training epochs were set to 20000, 100000 and 100000, respectively. If the loss did not decrease by more than 0.001 for 1000 consecutive epochs during the training process, the training was terminated early. To fully unleash the performance of CGBP, we applied a mixed strategy to adjust the hyperparameters. Specifically, we used the hyperopt[65] package to find the optimal combination of hyperparameters, which samples from the hyperparameter range set given by the user (50~300 maximum sampling times according to the difficulty of the problem). The dropout probability ranged from 0 to 0.5, and the learning rate ranged from 0.000001 to 0.1. In addition, for the hyperparameters $z$ and $\beta$ introduced in CSBP, each layer of the GNN model shared a separate initial chaotic strength $z$, where the two GNN layers were set to 3 and 1, respectively, and the initial embedding layer was set to 20. $\beta$ was set to 0.999 unless otherwise specified. The remaining unspecified parameters were set to the same as PI-GNN or the default value in Pytorch. All numerical experiments were performed on two Intel Xeon Gold 6226R CPUs and one Nvidia A100 GPU.

## Data availability

Random *d*-regular graphs were generated using the open-source networkx library (https://networkx.org). The benchmark datasets used in this work are all publicly available. The Gset and Queen datasets can be downloaded separately from https://web.stanford.edu/~yyye/yyye/Gset/ and https://mat.tepper.cmu.edu/COLOR02/. The three instances of the Citation dataset can be obtained from the references in the main text, or can be downloaded directly using the DGL library.

## Code availability

The Python code used in this study is available at https://github.com/PengTao-HUST/CGBP.

## Acknowledgments

This work was supported by the National Basic Research Program of China (No. 2022YFA1004800), Strategic Priority Research Program of the Chinese Academy of Sciences 418 (No. XDB38040400), National Natural Science Foundation of China (Nos. 31930022, 12131020, T2341007), the Special Fund for Science and Technology Innovation Strategy of Guangdong Province (Nos. 2021B0909050004, 2021B0909060002), JST Moonshot R&D (No. JPMJMS2021), the International Research Center for Neurointelligence (WPI-IRCN) at the University of Tokyo Institutes for Advanced Study (UTIAS), AMED under Grant Number JP22dm0307009, Institute of AI and Beyond of UTokyo, and JSPS KAKENHI Grant Number JP20H05921

# Author contributions

P.T. and L.N.C. conceived the idea. P.T. designed and performed the research. All authors analyzed and wrote the paper.

# Competing interests

The authors declare that they have no competing interests.

# References

1. Bengio Y, Lodi A, Prouvost A. Machine learning for combinatorial optimization: a methodological tour d'horizon. *European Journal of Operational Research* **290**, 405-421 (2021)
2. Naseri G, Koffas MA. Application of combinatorial optimization strategies in synthetic biology. *Nature communications* **11**, 2446 (2020)
3. Leppek K*, et al.* Combinatorial optimization of mRNA structure, stability, and translation for RNA-based therapeutics. *Nature Communications* **13**, 1536 (2022)
4. Feng L*, et al.* Explicit Evolutionary Multitasking for Combinatorial Optimization: A Case Study on Capacitated Vehicle Routing Problem. *IEEE Transactions on Cybernetics* **51**, 3143-3156 (2021)
5. Mazyavkina N, Sviridov S, Ivanov S, Burnaev E. Reinforcement learning for combinatorial optimization: A survey. *Computers & Operations Research* **134**, 105400 (2021)
6. Vinyals O, Fortunato M, Jaitly N. Pointer networks. In *Advances in neural information processing systems*, (2015)
7. Karalias N, Loukas A. Erdos goes neural: an unsupervised learning framework for combinatorial optimization on graphs. In *Advances in Neural Information Processing Systems*, 6659-6672 (2020)
8. Bello I, Pham H, Le QV, Norouzi M, Bengio S. Neural combinatorial optimization with reinforcement learning. *arXiv preprint arXiv:161109940*, (2016)
9. Khalil E, Dai H, Zhang Y, Dilkina B, Song L. Learning combinatorial optimization algorithms over graphs. In *Advances in neural information processing systems*, (2017)
10. Xin L, Song W, Cao Z, Zhang J. NeuroLKH: Combining deep learning model with Lin-Kernighan-Helsgaun heuristic for solving the traveling salesman problem. *Advances in Neural Information Processing Systems*, 7472-7483 (2021)
11. Helsgaun K. An effective implementation of the Lin–Kernighan traveling salesman heuristic. *European journal of operational research* **126**, 106-130 (2000)
12. Schuetz MJA, Brubaker JK, Katzgraber HG. Combinatorial optimization with physics-inspired graph neural networks. *Nature Machine Intelligence* **4**, 367-377 (2022)
13. Kingma DP, Ba J. Adam: A method for stochastic optimization. *arXiv preprint arXiv:14126980*, (2014)
14. Strogatz SH. *Nonlinear dynamics and chaos with student solutions manual: With applications to physics, biology, chemistry, and engineering*. CRC press (2018).
15. Chen L, Aihara K. Global searching ability of chaotic neural networks. *IEEE Transactions on Circuits and Systems I: Regular Papers* **46**, 974-993 (1999)
16. Goto H, Tatsumura K, Dixon AR. Combinatorial optimization by simulating adiabatic bifurcations in nonlinear Hamiltonian systems. *Science Advances* **5**, eaav2372 (2019)
17. Goto H*, et al.* High-performance combinatorial optimization based on classical mechanics. *Science Advances* **7**, eabe7953 (2021)
18. Chen L, Aihara K. Chaotic simulated annealing by a neural network model with transient chaos. *Neural networks* **8**, 915-930 (1995)
19. Chen L, Aihara K. Chaos and asymptotical stability in discrete-time neural networks. *Physica D* **104**, 286-325 (1997)
20. Guevara MR, Glass L, Shrier A. Phase locking, period-doubling bifurcations, and irregular dynamics in periodically stimulated cardiac cells. *Science* **214**, 1350-1353 (1981)
21. Kaplan DT, Clay JR, Manning T, Glass L, Guevara MR, Shrier A. Subthreshold dynamics in periodically stimulated squid giant axons. *Physical Review Letters* **76**, 4074-4077 (1996)
22. Aihara K, Numajiri T, Matsumoto G, Kotani M. Structures of attractors in periodically forced neural oscillators. *Physics Letters A* **116**, 313-317 (1986)
23. Skarda CA, Freeman WJ. How brains make chaos in order to make sense of the world. *Behavioral and Brain Sciences* **10**, 161-173 (1987)
24. Pool R. Is it healthy to be chaotic? *Science* **243**, 604-607 (1989)
25. Heltberg ML, Krishna S, Jensen MH. On chaotic dynamics in transcription factors and the associated effects in differential gene regulation. *Nature Communications* **10**, 71 (2019)
26. Destexhe A. Oscillations, complex spatiotemporal behavior, and information transport in networks of excitatory and inhibitory neurons. *Physical Review E* **50**, 1594 (1994)
27. Babloyantz A, Salazar JM, Nicolis C. Evidence of chaotic dynamics of brain

activity during the sleep cycle. *Physics Letters A* **111**, 152-156 (1985)

28. Matsumoto G, Aihara K, Hanyu Y, Takahashi N, Yoshizawa S, Nagumo J-i. Chaos and phase locking in normal squid axons. *Physics Letters A* **123**, 162-166 (1987)
29. Fontenele AJ, *et al.* Criticality between cortical states. *Physical Review Letters* **122**, 208101 (2019)
30. Fosque LJ, Williams-García RV, Beggs JM, Ortiz G. Evidence for quasicritical brain dynamics. *Physical Review Letters* **126**, 098101 (2021)
31. Marotto FR. On redefining a snap-back repeller. *Chaos, Solitons & Fractals* **25**, 25-28 (2005)
32. Lucas A. Ising formulations of many NP problems. *Frontiers in physics* **2**, 5 (2014)
33. Boettcher S. Analysis of the relation between quadratic unconstrained binary optimization and the spin-glass ground-state problem. *Physical Review Research* **1**, 033142 (2019)
34. Kipf TN, Welling M. Semi-supervised classification with graph convolutional networks. In *International Conference on Learning Representations*, (2017)
35. Hamilton W, Ying Z, Leskovec J. Inductive representation learning on large graphs. *Advances in neural information processing systems*, (2017)
36. Veličković P, Cucurull G, Casanova A, Romero A, Lio P, Bengio Y. Graph attention networks. *arXiv preprint arXiv:171010903*, (2017)
37. Tao P, Cheng J, Chen L. Brain-inspired chaotic backpropagation for MLP. *Neural Networks* **155**, 1-13 (2022)
38. Nagumo J, Sato S. On a response characteristic of a mathematical neuron model. *Kybernetik* **10**, 155-164 (1972)
39. Aihara K, Takabe T, Toyoda M. Chaotic neural networks. *Physics Letters A* **144**, 333-340 (1990)
40. Sutskever I, Martens J, Dahl G, Hinton G. On the importance of initialization and momentum in deep learning. *Proceedings of the 30th International Conference on Machine Learning* **28**, 1139-1147 (2013)
41. Amir D, Andrea M, Subhabrata S. Extremal cuts of sparse random graphs. *The Annals of Probability* **45**, 1190-1217 (2017)
42. Sherrington D, Kirkpatrick S. Solvable model of a spin-glass. *Physical review letters* **35**, 1792 (1975)
43. Goemans MX, Williamson DP. Improved approximation algorithms for maximum cut and satisfiability problems using semidefinite programming. *Journal of the ACM (JACM)* **42**, 1115-1145 (1995)
44. Benlic U, Hao J-K. Breakout local search for the max-cutproblem. *Engineering Applications of Artificial Intelligence* **26**, 1162-1173 (2013)
45. Choi C, Ye Y. Solving sparse semidefinite programs using the dual scaling algorithm with an iterative solver. *Manuscript, Department of Management Sciences, University of Iowa, Iowa City, IA* **52242**, (2000)

46. Kochenberger GA, Hao J-K, Lü Z, Wang H, Glover F. Solving large scale max cut problems via tabu search. *Journal of Heuristics* **19**, 565-571 (2013)

47. Toenshoff J, Ritzert M, Wolf H, Grohe M. Graph neural networks for maximum constraint satisfaction. *Frontiers in artificial intelligence* **3**, 580607 (2021)

48. Schuetz MJA, Brubaker JK, Zhu Z, Katzgraber HG. Graph coloring with physics-inspired graph neural networks. *Physical Review Research* **4**, 043131 (2022)

49. McCallum AK, Nigam K, Rennie J, Seymore K. Automating the construction of internet portals with machine learning. *Information Retrieval* **3**, 127-163 (2000)

50. Sen P, Namata G, Bilgic M, Getoor L, Galligher B, Eliassi-Rad T. Collective classification in network data. *AI magazine* **29**, 93-93 (2008)

51. Namata G, London B, Getoor L, Huang B, Edu U. Query-driven active surveying for collective classification. In *10th international workshop on mining and learning with graphs*, 1 (2012)

52. Hertz A, Werra Dd. Using tabu search techniques for graph coloring. *Computing* **39**, 345-351 (1987)

53. Li W, Li R, Ma Y, Chan SO, Pan D, Yu B. Rethinking Graph Neural Networks for the Graph Coloring Problem. *ArXiv* abs/2208.06975, (2022)

54. Fazayeli F, Wang L, Liu W. Back-propagation with chaos. *International Conference on Neural Networks and Signal Processing*, 5-8 (2008)

55. Zhang T, Cheng X, Jia S, Poo M-m, Zeng Y, Xu B. Self-backpropagation of synaptic modifications elevates the efficiency of spiking and artificial neural networks. *Science Advances* **7**, eabh0146 (2021)

56. Hinton G. The forward-forward algorithm: Some preliminary investigations. *arXiv preprint arXiv:221213345*, (2022)

57. Zhuang J*, et al.* Adabelief optimizer: Adapting stepsizes by the belief in observed gradients. *Advances in neural information processing systems*, 18795-18806 (2020)

58. Xie X, Zhou P, Li H, Lin Z, Yan S. Adan: Adaptive Nesterov Momentum Algorithm for Faster Optimizing Deep Models. *arXiv preprint arXiv:220806677*, (2022)

59. Ye Y, Ji S. Sparse graph attention networks. *IEEE Transactions on Knowledge and Data Engineering* **35**, 905-916 (2021)

60. Wang MY. Deep graph library: Towards efficient and scalable deep learning on graphs. In *ICLR workshop on representation learning on graphs and manifolds*, (2019)

61. Nair V, Hinton GE. Rectified linear units improve restricted boltzmann machines. *International Conference on Machine Learning*, 807-814 (2010)

62. Srivastava N, Hinton G, Krizhevsky A, Sutskever I, Salakhutdinov R. Dropout: a simple way to prevent neural networks from overfitting. *The journal of*

*machine learning research* **15**, 1929-1958 (2014)

63. Ioffe S, Szegedy C. Batch normalization: Accelerating deep network training by reducing internal covariate shift. In *Proceedings of the 32nd International Conference on Machine Learning*, 448-456 (2015)
64. Paszke A*, et al.* Pytorch: An imperative style, high-performance deep learning library. *Advances in neural information processing systems* **32**, 8026-8037 (2019)
65. Bergstra J, Yamins D, Cox D. Making a science of model search: Hyperparameter optimization in hundreds of dimensions for vision architectures. In *International conference on machine learning*, 115-123 (2013)